\newcommand{\CLASSINPUTtoptextmargin}{0.85 in}
\newcommand{\CLASSINPUTbottomtextmargin}{1.1 in}

\documentclass[conference]{IEEEtran}
\IEEEoverridecommandlockouts
\usepackage{cite}
\usepackage{amsmath,amssymb,amsfonts}
\usepackage{graphicx}
\usepackage{textcomp}
\usepackage{xcolor}
\usepackage{mathtools}
\usepackage{xcolor}
\usepackage{tabu}
\usepackage{multirow}
\usepackage{tabularx}
\usepackage{tabularray}
\usepackage{subfig}
\usepackage{graphicx}
\usepackage[english]{babel}
\usepackage{pifont}
\usepackage{roboto-mono}    % roboto for the typewriter font
\usepackage[T1]{fontenc}    % use 8-bit T1 fonts

\usepackage{tikz}
\newcommand{\ballnumber}[1]{\tikz[baseline=(myanchor.base)] \node[circle,fill=.,inner sep=1pt] (myanchor) {\color{-.}\bfseries\footnotesize #1};}

\usepackage{enumitem}
\newlist{inlineroman}{enumerate*}{1}
\setlist[inlineroman]{itemjoin*={{, and }},afterlabel=~,label=\roman*.}

\usepackage{hyperref}

\usepackage[scaled]{beramono}
\usepackage{listings}
\usepackage{multirow}
\lstdefinestyle{mystyle}{
    language=Python,
    basicstyle=\ttfamily\footnotesize,
    keywordstyle=\bfseries,
    commentstyle=\color{gray},
    numberstyle=\ttfamily\color{gray},
    numbers=left,
    xleftmargin=2.5em,
    mathescape=true,
    captionpos=b,
}
\usepackage[linesnumbered,ruled]{algorithm2e}

\usepackage{xspace}

\newcommand{\rlpolicy}{\texttt{RL-base}\xspace}
\newcommand{\rlpolicyhybrid}{\texttt{RL-Hybrid}\xspace}

\makeatletter
\newcommand{\linebreakand}{%
  \end{@IEEEauthorhalign}
  \hfill\mbox{}\par
  \mbox{}\hfill\begin{@IEEEauthorhalign}
}
\makeatother

\newcommand{\contentionsensitivity}{CS}
\newcommand{\util}{Util}
\newcommand{\throughputideal}{Throughput_{\text{ideal}}}
\newcommand{\throughputcontention}{Throughput_{\text{contention}}}
\newcommand{\wone}{W_{1}}
\newcommand{\wtwo}{W_{2}}
\def\BibTeX{{\rm B\kern-.05em{\sc i\kern-.025em b}\kern-.08em
    T\kern-.1667em\lower.7ex\hbox{E}\kern-.125emX}}
\begin{document}

\title{Network Contention-Aware Cluster Scheduling with Reinforcement Learning}

\author{
\IEEEauthorblockN{Junyeol Ryu}
\IEEEauthorblockA{\textit{Computer Science and Engineering} \\
\textit{Seoul National University}\\
Seoul, Korea \\
junyeol@aces.snu.ac.kr}
\and
\IEEEauthorblockN{Jeongyoon Eo}
\IEEEauthorblockA{\textit{Computer Science and Engineering} \\
\textit{Seoul National University}\\
Seoul, Korea \\
jeongyoon.eo@snu.ac.kr}
}

\maketitle

\begin{abstract}
With continuous advances in deep learning, distributed training is becoming common in GPU clusters. Specifically, for emerging workloads with diverse amounts, ratios, and patterns of communication, we observe that network contention can significantly degrade training throughput. However, widely used scheduling policies often face limitations as they are agnostic to network contention between jobs.
In this paper, we present a new approach to mitigate network contention in GPU clusters using reinforcement learning. We formulate GPU cluster scheduling as a reinforcement learning problem and opt to learn a network contention-aware scheduling policy that efficiently captures contention sensitivities and dynamically adapts scheduling decisions through continuous evaluation and improvement. We show that compared to widely used scheduling policies, our approach reduces average job completion time by up to 18.2\% and effectively cuts the tail job completion time by up to 20.7\% while allowing a preferable trade-off between average job completion time and resource utilization. 
\end{abstract}


\section{Introduction}
\label{sec:intro}

Distributed deep learning (DL) training is becoming increasingly prevalent in GPU clusters. A recent analysis published by Alibaba~\cite{mlaas} has reported that over 80\% of the total submitted DL training jobs\footnote{Combination of DL model to train, GPU demand, and scheduled nodes.} run on multiple GPUs spanning over multiple nodes. This rate corresponds to approximately 5$\times$ increase in five years from the previous report from Microsoft~\cite{philly}. 
Also, emerging DL training workloads present diverse amounts, ratios, and patterns of communication. For instance, Fully Sharded Data-Parallel training (FSDP)~\cite{fairscale-fsdp,pytorch-fsdp} features heavy communication, with at least 50\% increased communication cost compared to conventional data-parallel training~\cite{zero}. Mixture of Experts (MoE)~\cite{moe} training implements a gating network-based routing mechanism between distributed experts using {\texttt{AllToAll}\xspace} pattern that has higher communication cost than traditional {\texttt{AllReduce}\xspace} pattern.
Eventually, previous studies have shown that GPU clusters can suffer significant performance degradation due to conflicts in network communication when distributed training and emerging workloads with diverse communication are common~\cite{muri,mlaas,rajasekaran2022congestion}.
Ultimately, this trend raises a new challenge to GPU cluster scheduling: mitigating performance slowdown due to \textit{network contention}.

Notwithstanding the new challenge, we notice that widely used scheduling policies (e.g., LAS~\cite{tiresias,las} and SRTF~\cite{srtf}) are often agnostic to network contention between jobs, hence are prone to significant degradation in jobs' training throughput. 
For example, when FSDP and MoE training share networks, they suffer up to 49.1\% and 66.7\% throughput degradation compared to isolated cluster, respectively. 
Unfortunately, such unfavorable scheduling decisions cannot be avoided under contention-agnostic policies. 
Yet, we observe that a job experiences varying degrees of network contention based on the model and the placement of the co-located\footnote{Scheduling jobs such that they become each other's node-sharing neighbors, sharing all or part of the allocated nodes, and thus bandwidth of intra and inter-node networks.} jobs. 
Accordingly, the throughput degradation of the aforementioned example can be improved up to 21.6\% and 19.5\%, respectively, depending on how they are co-located.
Building upon this insight, we define \textit{contention sensitivity} ($\contentionsensitivity$) of a job as the ratio of its ideal throughput to the degraded throughput when co-located with another job (\ref{eq:1}). 
%This definition establishes contention sensitivity as the fundamental metric for network contention.
\begin{equation}
    \label{eq:1}
    \contentionsensitivity = \frac{\throughputideal}{\throughputcontention}
\end{equation}

% requirement
In this paper, we propose a method for scheduling distributed DL jobs in GPU clusters that minimizes network contention. Two notable challenges are as follows: 1. efficiently capturing contention sensitivities, and 2. dynamically adapting to diverse distributions of jobs and their contention sensitivities. To address these challenges, we devise a reinforcement learning (RL)-based approach to swiftly learn an effective scheduling policy by continuous evaluation and improvement of scheduling decisions across diverse distributions of jobs.
Specifically, we make the following contributions:
\begin{itemize}
    \item We propose a novel design that translates the network contention problem in cluster scheduling into an RL problem. We show that our design can efficiently capture contention sensitivities of jobs and dynamically adapt scheduling decisions across diverse distributions of jobs.
    \item We present an end-to-end system for training scheduling policies with RL to its deployment on GPU clusters. We provide two initial scheduling policies, \rlpolicy and \rlpolicyhybrid, and implement mechanisms that execute the decisions of the scheduling policy. 
    \item We evaluate our scheduling policies with a variety of DL training job traces on a GPU cluster. \rlpolicy outperforms LAS and SRTF by reducing average JCT by up to 18.2\% and effectively cutting tail JCT by up to 20.7\%, while \rlpolicyhybrid achieves a preferable trade-off between average JCT and resource utilization. 
    \item We open our work at (\url{https://github.com/gajagajago/deepshare}) as a community asset for future research in RL-based GPU cluster scheduling.
\end{itemize}
\section{Background}
\label{sec:back}

\noindent
\textbf{DL training in GPU clusters.}
%intro
\textit{Deep learning} is a process to train a deep neural network (DNN) with the purpose of analyzing and deriving valuable knowledge from data~\cite{shinde2018review}. The DNN is constructed with numerous layers and parameters. During training, the DNN is tasked with making predictions and updating its parameters based on the computation of errors in relation to actual outcomes. Given its inherent computational intensity, training is usually performed on high-capacity accelerators like GPUs. Due to extremely high price (e.g., NVIDIA A100 costs about \$10k), organizations often build GPU clusters to be shared by users and production groups~\cite{synergy}. Consequently, GPU clusters usually employ a scheduler for the efficient allocation of cluster resources. 
These schedulers predominantly operate under two key objectives: reducing Job Completion Time (JCT)~\cite{optimus,tiresias,gavel,afs,pollux,muri} and enhancing resource utilization~\cite{gandiva,tiresias,antman,synergy}. Thus, these two performance metrics form the foundation of our RL formulation's reward (Section \ref{sec:rl}).

\noindent
\textbf{RL for cluster scheduling.} 
% General intro to RL
\textit{Reinforcement learning} involves an agent that learns to make better decisions directly from experience interacting with the environment~\cite{sutton2018reinforcement}. 
The agent learns by \textit{reinforcement}, wherein it receives rewards contingent on the quality of its decisions~\cite{deeprm}. 
RL has recently become an active area of research in machine learning~\cite{mnih2016asynchronous,mnih2013playing,mnih2015human,schulman2015trust,silver2016mastering}, and RL-based approaches have demonstrated great potential in various domains including congestion control~\cite{pcc-rl}, video streaming~\cite{pensieve, comyco, merina}, real-time communication~\cite{r3net, concerto, onrl, loki, hrcc}, and resource management~\cite{sibyl,deeprm,deepplace-apsys,deepplace-aaai}.
% Benefits of RL
RL approaches are known to be especially well-suited to resource management systems due to the followings:
\begin{itemize}
    \item Decisions made by the systems are highly repetitive, leaving abound of training data (e.g., scheduling decisions and corresponding effects) to the RL algorithm.
    \item Reward can reflect complex objectives (e.g., JCT and utilization) which are difficult to model analytically.
    \item Agent are highly adaptive to constantly shifting or even previously unseen circumstances.
\end{itemize}

% prev works
DeepRM~\cite{deeprm} and DeepPlace~\cite{deepplace-apsys,deepplace-aaai} are two notable works that apply RL in resource management. 
DeepRM represents the system state as a combination of cluster resources and resource demands of jobs. In this scheme, cluster resources are equipped with available time slots for job allocation, while jobs demand distinct quantities of time slots for different resource types. The action step revolves around matching jobs' time slots with those of resources, and the reward is designed to mirror the objective of minimizing average slowdown.
DeepPlace builds upon DeepRM and introduces a more intricate reward that incorporates a blend of purpose-oriented cues, including factors like \textit{resource contention penalty} and \textit{under-utilization penalty}. 

% motiv
Nevertheless, there exists a limitation when attempting to employ these methodologies in scheduling distributed DL jobs in GPU clusters.
First, since these approaches primarily focus on host resources (such as CPU and DRAM) and address small containerized microservices as their target jobs, their state representation is inadequate for capturing distributed GPU jobs.
Second, their action scope remains limited to within a single node, making it impractical to schedule distributed jobs spanning multiple nodes.
Finally, due to the absence of consideration for network contention between jobs, their reward structure fails to encompass the performance impact of network contention.
% motivation
Therefore, our main objective is to introduce an efficient RL-based solution for GPU cluster scheduling that efficiently handles network contention problem (Section \ref{sec:motiv}).

\section{Motivation}
\label{sec:motiv}

\definecolor{MineShaft}{rgb}{0.2,0.2,0.2}
\begin{table*}[t]
% \vspace{3mm} % for top margin warning of edas
\centering
\begin{tblr}{
  row{6} = {r},
  row{7} = {r},
  cell{1}{2} = {r},
  cell{1}{3} = {r},
  cell{1}{4} = {r},
  cell{1}{5} = {r},
  cell{1}{6} = {r},
  cell{1}{7} = {r},
  cell{2}{2} = {r},
  cell{2}{3} = {r},
  cell{2}{4} = {r},
  cell{2}{5} = {r},
  cell{2}{6} = {r},
  cell{2}{7} = {r},
  cell{3}{2} = {r},
  cell{3}{3} = {r},
  cell{3}{4} = {r},
  cell{3}{5} = {r},
  cell{3}{6} = {r},
  cell{3}{7} = {r},
  cell{4}{2} = {r},
  cell{4}{3} = {r},
  cell{4}{4} = {r},
  cell{4}{5} = {r},
  cell{4}{6} = {r},
  cell{4}{7} = {r},
  cell{5}{1} = {r=3}{},
  cell{5}{2} = {r},
  cell{5}{3} = {r},
  cell{5}{4} = {r},
  cell{5}{5} = {r},
  cell{5}{6} = {r},
  cell{5}{7} = {r},
  cell{6}{2} = {fg=MineShaft},
  cell{7}{2} = {fg=MineShaft},
  hline{1-5,8} = {-}{},
}
Task           & Model (Abbreviation)              & Dataset      & {Trainable \\Parameters} & {Average Bandwidth \\Consumption (MB/s)} & {$\frac{Comm}{Comp}$} & {Communication Patterns} \\
Graph & GraphSage (GNN)~\cite{graphsage} & Reddit & 0.33M & 24.63 & 0.57 & {\texttt{AllReduce}\xspace} \\
Image & MobileNetV3 (IMG)~\cite{mobilenet} & ImageNet & 2.04M & 211.25 & 2.43 & {\texttt{AllReduce}\xspace} \\
Recommendation & DLRM (DLRM)~\cite{dlrm} & Criteo & 333.32M & 170.28 & 13.36 & {\texttt{AllReduce}\xspace} \\
Language & Transformer-XL (LM)~\cite{transformerxl} & Wikitext-103 & 202.44M & 854.82     & 1.87 & {\texttt{AllReduce}\xspace} \\
& GPT-2 (FSDP)~\cite{radford2019language} & Wikitext-2 & 184.89M & 2672.40 & 7.32 & {\texttt{ReduceScatter}\xspace, \texttt{AllGather}\xspace} \\
& GPT-2 (MoE)~\cite{radford2019language}  & Wikitext-2  & 268.92M & 929.48 & 13.79 & {\texttt{AllToAll}\xspace}
\end{tblr}
\caption{Models used in this work. $\frac{Comm}{Comp}$ denotes communication-to-computation ratio. Average bandwidth consumption and $\frac{Comm}{Comp}$ is profiled in our evaluation environment (Section \ref{sec:eval}).}
\label{table:models}
\end{table*}

In this section, we introduce emerging DL training workloads encompassing diverse communication characteristics. Then, we analyze their contention sensitivities to draw insights for effective network contention-aware cluster scheduling. We use these workloads (and their variations) to constitute the jobs traces for evaluation (Section \ref{sec:eval}).

\noindent
\textbf{Communication characteristics of emerging DL training workloads (\autoref{table:models}).} 
FSDP~\cite{pytorch-fsdp,fairscale-fsdp} and MoE~\cite{moe} exhibit high network bandwidth consumption and communication-to-computation ratio. Specifically, FSDP and MoE demonstrate respectively 3.01$\times$ and 5.67$\times$ higher communication-to-computation ratio and 12.65$\times$ and 4.40$\times$ larger average bandwidth consumption compared to a traditional image model training such as MobileNetV3~\cite{mobilenet}. These characteristics stem from their communication pattern, designed to achieve GPU memory efficiency at the cost of increased communication. Concretely, FSDP adds per-layer {\texttt{AllGather}\xspace} at forward pass, and per-layer {\texttt{AllGather}\xspace} and {\texttt{ReduceScatter}\xspace} at backward pass compared to conventional data-parallel training~\cite{pytorch-fsdp, fairscale-fsdp}. MoE adds gated network-based {\texttt{AllToAll}\xspace} communication between distributed experts~\cite{moe}.

DLRM~\cite{dlrm} and Transformer-XL~\cite{transformerxl} display either a high communication-to-computation ratio or high network bandwidth consumption. Especially, DLRM exhibits a 5.49$\times$ higher communication-to-computation ratio but 19.3\% lower bandwidth consumption compared to MobileNetV3. Collective communications account for a significant fraction of time in training DLRM at scale~\cite{yangdlrmkdd}. This is because recommendation tasks such as DLRM spend around 80\% of total training time on host resources~\cite{mlaas} as sparse computation of element-wise operators dominates. On the other hand, Transformer-XL exhibits a 23.04\% lower communication-to-computation ratio but 4.04$\times$ higher bandwidth consumption compared to MobileNetV3. This is because the Transformer~\cite{vaswani2017attention} blocks employ the attention mechanism that requires larger amounts of computation compared to the convolution-based mechanism of image models. 

GraphSage~\cite{graphsage} exhibits the lowest network bandwidth consumption and communication-to-computation ratio, with 76.85\% less communication-to-computation ratio and 88.34\% lower average bandwidth consumption compared to MobileNetV3. This is because GraphSage training partitions the input graph among nodes, where per-node graph preprocessing takes 30-90\% of the training time with involving only a little communication~\cite{mlaas}. 

\begin{figure}[t]
    \centering
    \includegraphics[width=0.95\columnwidth]{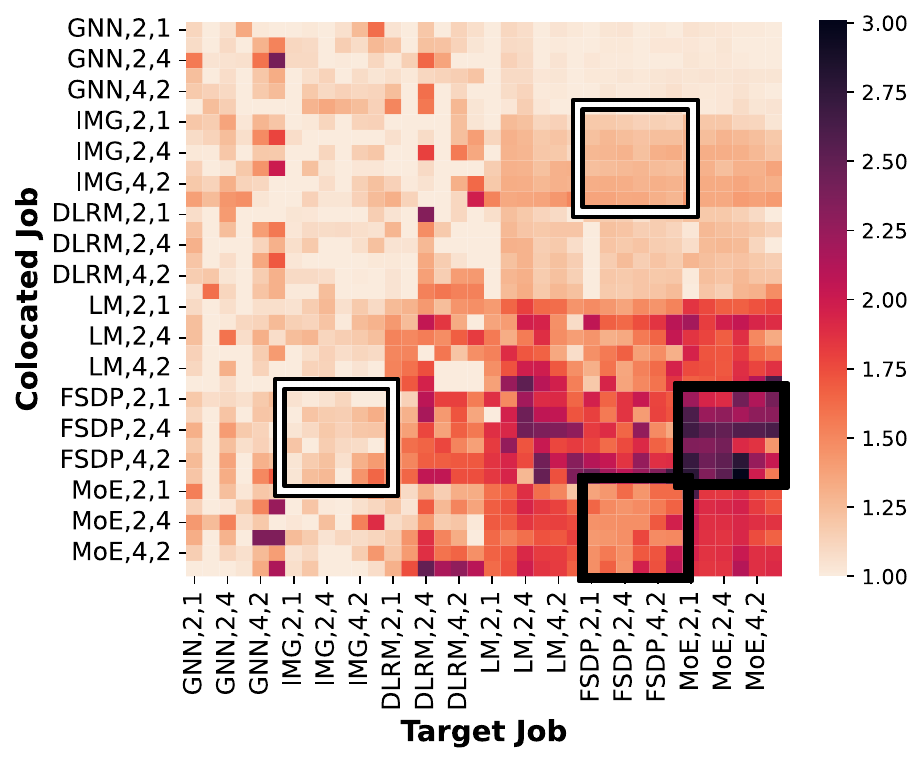}
    % \vspace{-3mm}
    \caption{Contention sensitivity heat map. Darker value indicates higher contention sensitivity, which signifies larger throughput degradation of the target job according to the network contention from the co-located job. Numbers behind the model name denote nodes and GPUs per node, respectively.}
    \label{fig:motiv-heatmap}
    % \vspace{-5mm}
\end{figure}

\noindent
\textbf{Contention sensitivity of a job varies by its co-located job.} 
In \autoref{fig:motiv-heatmap}, each pixel value depicts the contention sensitivity of the target job according to the network contention from the co-located job. A pair of a target model and a co-located model is represented as a $6\times6$ grid (e.g. the black boxes in \autoref{fig:motiv-heatmap} shows the varying contention sensitivities of FSDP and MoE, with respect to diverse GPU demands and node assignments). 
%ex
We observe that some jobs (GNN and IMG) show consistent contention sensitivities regardless of co-located jobs, whereas the others (DLRM, LM, FSDP, and MoE) show high variability with regard to the model, GPU demands, and node assignment of the co-located job. For example, when FSDP and MoE training are co-located, they experience contention sensitivities of up to 1.96 and 3.00, respectively, with varying degrees according to their co-location (the black boxes in \autoref{fig:motiv-heatmap}). In contrast, when FSDP and MobileNetV3 training are co-located, they exhibit moderate degrees of contention sensitivities with at most 1.35 and 1.43, respectively (the white boxes in \autoref{fig:motiv-heatmap}).

We summarize our findings from this section as follows:
\begin{itemize}
    \item Jobs exhibit a variety of communication characteristics, which contribute to their varying contention sensitivities when co-located with other jobs.
    \item In this regard, scheduling based on efficiently captured contention sensitivities and aimed at reducing expected contention will effectively alleviate network contention.
\end{itemize}          
\section{RL Formulation}
\label{sec:rl}

\noindent
\textbf{Requirements.}
Summarizing the key insights from previous sections, the main requirements of an ideal network contention-aware GPU cluster scheduling includes: 
\begin{itemize}
    \item \textbf{R1.} Fast adaptation of its decisions to reflect the constantly changing distribution of contention sensitivities as scheduled jobs change.
    \item \textbf{R2.} Minimizing cluster-wide performance degradation due to contention, achieving low average and tail JCT while maintaining high resource utilization.
\end{itemize}

To satisfy the requirements, we propose our design for a network contention-aware scheduling with RL. We explain our formulation of state, action, reward, and training algorithm. We also present two initial scheduling policies (\textit{agents} in RL term) trained with RL, namely \rlpolicy and \rlpolicyhybrid. These policies serve as benchmarks for evaluation (Section \ref{sec:eval}).

\begin{figure}[t]
    \centering
    \includegraphics[width=0.99\columnwidth]{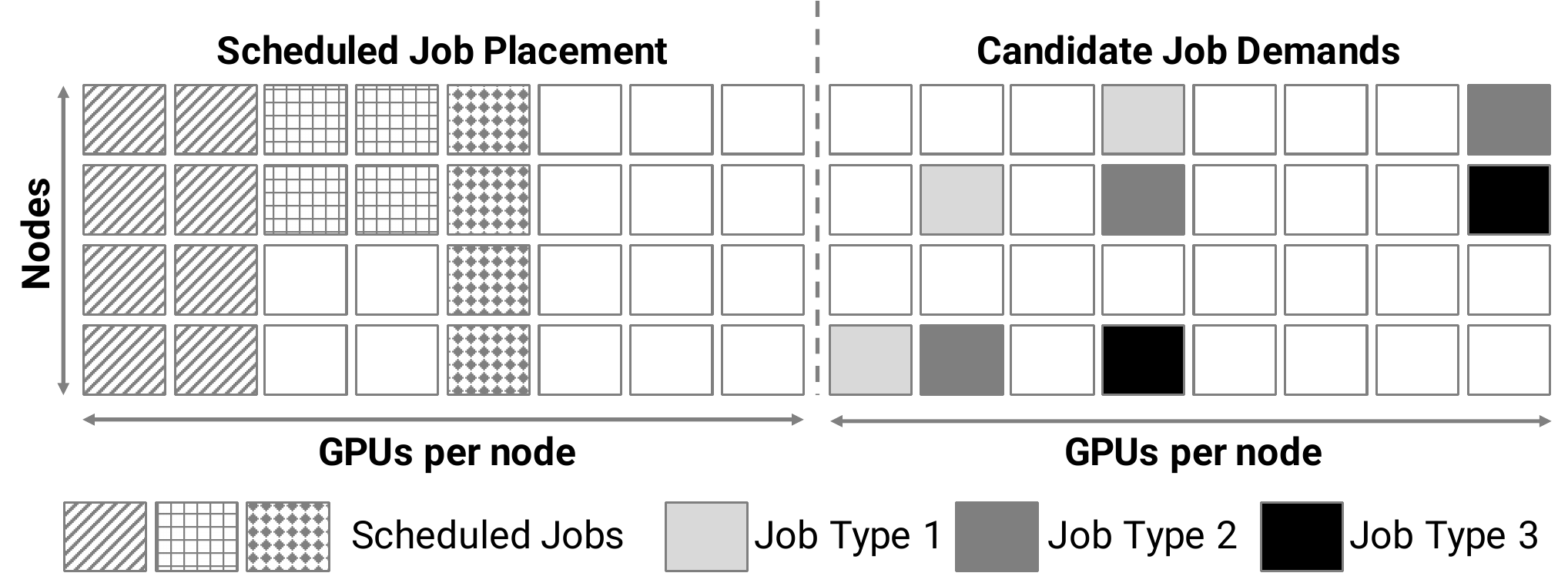}
    \caption{State representation. (Example for 4 nodes each with 8 GPUs, 3 scheduled jobs and 3 candidate jobs with varying in GPU demands)}
    \label{fig:state}
    % \vspace{-5mm}
\end{figure}

\noindent
\textbf{State.}
% intro
To adapt the agent's decisions such that it reflects the constantly changing contention sensitivities (\textbf{R1}), we develop a state representation that captures the cluster-wide co-location of jobs. \autoref{fig:state} illustrates an example of our state design. We represent the cluster state as a two-dimensional tensor of shape <Nodes, $2\times$GPUs per node>. This fixed state design satisfies a desirable attribute to be applied as input to neural network-based RL training algorithms. The left half encodes the physical placement of scheduled jobs. Each slot contains the profile of a job, if any job is scheduled on the GPU corresponding to the slot. Conversely, the right half captures the resource demands of candidate jobs in the waiting queue. For example, if a candidate job is encoded in $slot_{i,j}$ of the right half, it signifies that one feasible placement option for the job is distributing its total $j*2^i$ demands across $2^i$ nodes with $j$ GPUs each. Since there are various ways to partition a job's demand, jobs with multiple GPU demands can be encoded in multiple slots of the right half. 

\IncMargin{.2em} % DO NOT REMOVE!! EDAS CHECKER
\begin{algorithm}[t]
\DontPrintSemicolon
\SetNoFillComment
\SetAlgoLined
\caption{RL Training Algorithm\label{alg:rl-train-alg}}
\SetKwFunction{FMain}{Train}
\SetKwFunction{FR}{Round}
\SetKwFunction{FCR}{ComputeReward}

\SetKwProg{Fn}{Function}{:}{}
\SetKwProg{Pn}{Function}{:}{\KwRet}

\SetKwInOut{Parameter}{Parameter}
\SetKwInOut{Input}{Input}

\Input{$Trace$: job trace, $Alg$: RL algorithm}
\Parameter{$\wone,\wtwo$: weights, $T$: scheduling interval, $F$: policy checkpoint path}

$C \gets$ Init empty cluster environment \;
$Q \gets$ Waiting jobs in $Trace$ \;
$Policy \gets$ Init policy with $Alg$ \;
\While{!$C$.empty or !$Q$.empty} {\label{lst:line:round-while-loop-start}
    \FMain{$C$, $Q$, $Policy$} \;
} \label{lst:line:round-while-loop-end}
Save $Policy$ to $F$ \label{lst:line:ckpt}\;

\;
\Fn{\FR{$C$, $Q$, $Policy$}}{
    $State \gets$ Get state from $C$ and $Q$ \label{lst:line:get-state} \;
    $Actions \gets$ $Policy$ computes action from $State$ \label{lst:line:get-action}\;
    \For{$job, nodes \in Actions$}{\label{lst:line:sched-start}
        Schedule $job$ to $nodes$ \;
    } \label{lst:line:sched-end}
    Sleep for $T$ \label{lst:line:wait-interval}\;
    $Reward \gets$ \FCR{$C$} \label{lst:line:compute-reward}\;
    Update $Policy$ with $Reward$ \label{lst:line:get-reward}\;
}
\;
\Fn{\FCR{$C$}}{
    $\contentionsensitivity$, $\util \gets$ 0 \; 
    \For{$job \in C.jobs$}{
        $CS_{\text{job}} \gets$ Profile contention sensitivity of $job$\;
        Update $\contentionsensitivity$ with $CS_{\text{job}}$ \;
    }
    \For{$node \in C.nodes$} {
        \For{$GPU \in node$} {
            $Util_{\text{GPU}} \gets$ Profile utilization of $GPU$ \;
            Update $\util$ with $Util_{\text{GPU}}$ \;
        }
    }
    \KwRet $-\wone * \contentionsensitivity + \wtwo * \util$ \;
}
\end{algorithm}

\noindent
\textbf{Action.}
%intro
Choosing an arbitrary set of candidate jobs from the waiting queue can lead to a large action space of up to $2^Q$ when the queue length is $Q$.
To reduce the size of the action space for faster convergence of the training algorithm, $K$ candidate jobs with different GPU demands are picked starting from the head of the waiting queue (forming the right half of the state representation).
Besides, to keep an action space of constant size without invalid actions, our action space for the candidate jobs is a set of lists of nodes to schedule the targets. $\emptyset$ denotes that the agent decided not to schedule any jobs in the current round. 
%ex
For example, if the agent returns {\texttt{[Node1, Node2]}\xspace} as an action for job type 1 (\autoref{fig:state}), it is assigning the job to Node1 and Node2 with two GPUs each as job type 1 is encoded in $slot_{2,2}$ on the right half of the state representation.
% preempt
One possibly beneficial extension of the action space can be to migrate a job if there is a better placement option or preempt if the current co-location is detrimental. But for the current prototype, we only support preemption based on a predefined threshold on contention sensitivity and plan to incorporate this feature in the action space soon.

\noindent
\textbf{Reward.}
To minimize cluster-wide performance degradation due to contention (\textbf{R2}), our reward penalizes an increase in cluster-wide average contention sensitivity of scheduled jobs and provides incentives on higher cluster-wide average GPU utilization (\ref{eq:2}). The two terms articulate the reward because reduced average job contention sensitivity is positively related to reduced JCT, and high GPU utilization corresponds to enhanced resource utilization, which are the two pivotal performance metrics of GPU cluster schedulers (Section \ref{sec:back}). Weights ${\wone}$ and $\wtwo = 1 - \wone$ are imposed on contention sensitivity (\contentionsensitivity) and utilization (\util) term. Its values can be tailored to reflect the relative preference between average contention sensitivity and utilization, as the two terms typically have a trade-off relationship. For instance, co-locating as many jobs as possible to maximize resource utilization may exacerbate average contention sensitivity, leading to more network contention experienced by the co-located jobs.
\begin{equation}
    \label{eq:2}
    Reward = -\wone * \contentionsensitivity + \wtwo * \util 
\end{equation}

\noindent
\textbf{Training algorithm.}
Algorithm \ref{alg:rl-train-alg} illustrates the training algorithm. The agent ({\texttt{Policy}\xspace}) is trained with a neural network-based RL algorithm ({\texttt{Alg}\xspace}) using job traces ({\texttt{Trace}\xspace}) on a simulated GPU cluster environment. The agent makes scheduling decisions until all jobs have finished (Lines \ref{lst:line:round-while-loop-start}-\ref{lst:line:round-while-loop-end}). For every scheduling round, the agent is given the state representation (Line \ref{lst:line:get-state}) and chooses an action (Line \ref{lst:line:get-action}). Jobs are scheduled as depicted in the action (Lines \ref{lst:line:sched-start}-\ref{lst:line:sched-end}). After waiting for the scheduling interval (Line \ref{lst:line:wait-interval}), reward is computed with respect to contention sensitivities of the scheduled jobs and resource utilization of the cluster (Line \ref{lst:line:compute-reward}) and is returned to the agent to adapt its policy using the neural network-based RL algorithm ({\texttt{Alg}\xspace}) (Line \ref{lst:line:get-reward}). By iterating through numerous scheduling rounds in the training process, the agent continuously explores diverse job co-location options and adapts its scheduling decisions to effectively mitigate cluster-wide network contention. When the training finishes, the agent is saved into a file as a trained scheduling policy (Line \ref{lst:line:ckpt}).

\noindent
\textbf{Benchmark policies.} 
We present two initial scheduling policies trained with RL, namely \rlpolicy and \rlpolicyhybrid. These policies serve as benchmarks for evaluation (Section \ref{sec:eval}).
\begin{itemize}
    \item \rlpolicy makes scheduling decisions only based on the action of the trained policy. It may decide not to schedule jobs even when the cluster has enough resources to avoid an increase in performance degradation due to contention.
    \item \rlpolicyhybrid performs decision-level multiplexing of the trained policy and a simple rule-based policy. \footnote{Such hybrid design is in line with the recent trend of incorporating the decisions of rule-based policies to guide or aid the RL-based policy to achieve both the RL-based one's high adaptivity and the rule-based one's stability and interpretability~\cite{onrl, loki, hrcc, orca}.} It usually follows the decisions of the trained policy, but when it faces an $\emptyset$, i.e. decision not to schedule, it follows a safety rule of trying greedy scheduling to prevent low utilization.
\end{itemize}             
\section{System}
\label{sec:system}

\begin{figure}[t]
    \centering
    \includegraphics[width=.99\columnwidth]{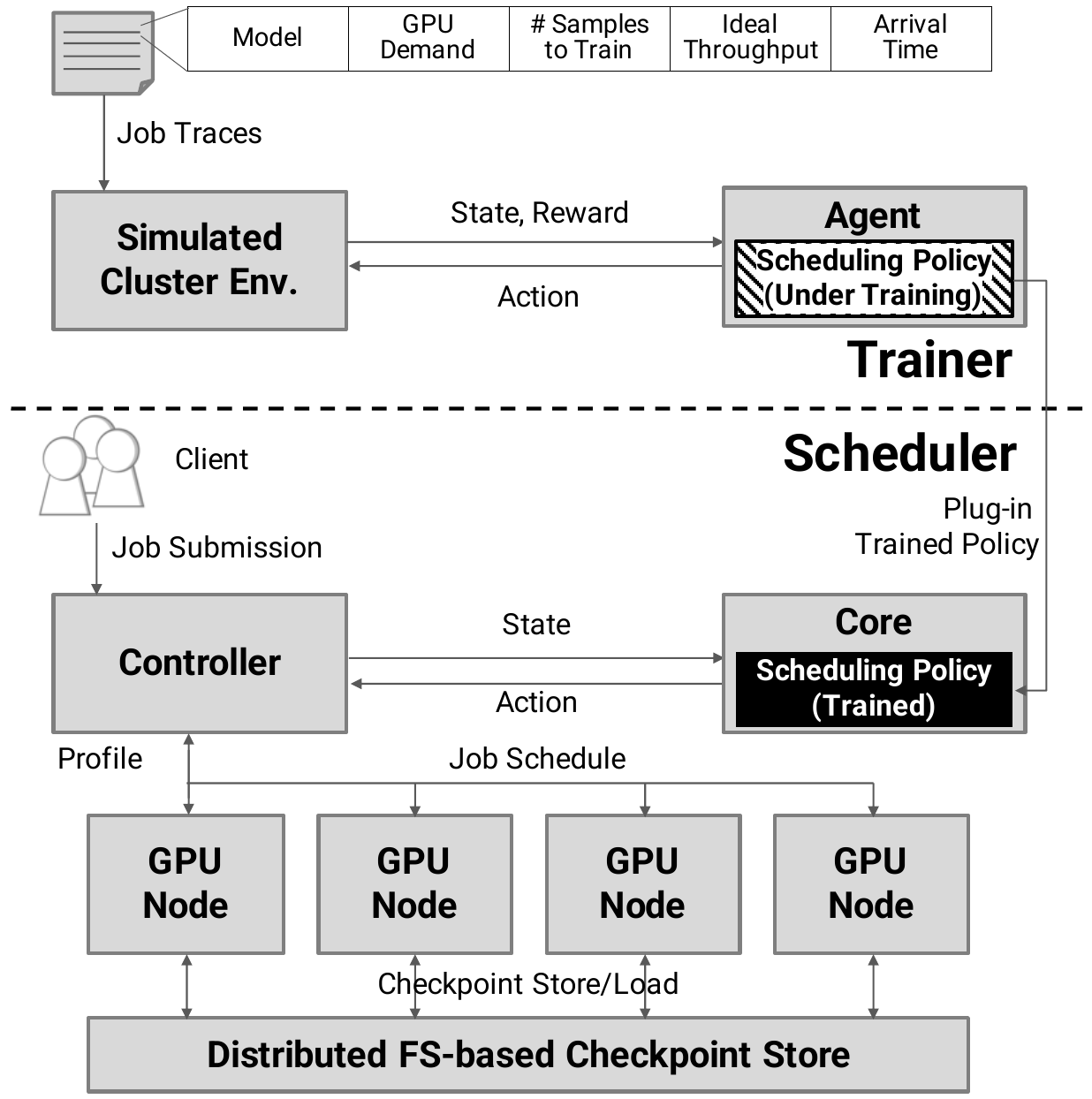}
    \caption{System overview. The policy is trained on a trace-driven simulated GPU cluster environment using RL and deployed on the actual GPU cluster.}
    \label{fig:system-overview}
    % \vspace{-1mm}
\end{figure}

\subsection{Framework Overview}
\label{sec:system-overview}

We build an end-to-end system from training a scheduling policy with RL on a simulated cluster environment to its deployment on GPU clusters. \autoref{fig:system-overview} illustrates the overall system architecture. The system is composed of two parts: {\texttt{Trainer}\xspace} and {\texttt{Scheduler}\xspace}. {\texttt{Trainer}\xspace} implements a simulated cluster environment and the training algorithm to train a scheduling policy with RL (Section \ref{sec:rl}). {\texttt{Scheduler}\xspace} deploys the trained policy on GPU clusters and provides mechanisms to execute the scheduling decisions.

\subsection{Trainer}
\label{sec:simulator}

{\texttt{Trainer}\xspace} provides OpenAI gym~\cite{brockman2016openai}-based simulated cluster environment of nodes and GPUs. It also provides a common interface for a set of off-the-shelf RL algorithms 
%(e.g., A2C~\cite{a2c}, PPO~\cite{ppo}, DQN~\cite{mnih2013playing}, TD3~\cite{td3}, and SAC~\cite{sac}) 
in widely used stable-baselines3~\cite{stable-baselines3}. Hence, researchers can easily customize the RL formulation and train their own policies. {\texttt{Trainer}\xspace} requires only job trace (e.g., Microsoft's Philly trace~\cite{philly} and Alibaba's PAI trace~\cite{mlaas}) to start training on simulated cluster environment. Training an agent on an episode of 256 jobs completes in less than a minute. 

\subsection{Scheduler}
\label{sec:scheduler}

{\texttt{Scheduler}\xspace} is implemented on top of Slurm~\cite{yoo2003slurm}, an open-sourced Linux cluster manager, and uses HDFS~\cite{shvachko2010hadoop} as the checkpoint store. {\texttt{Controller}\xspace} is implemented on top of Slurm control daemon ({\texttt{slurmctld}\xspace}) by adding global data structures for cluster state and waiting queues, an interface for communicating with the scheduler and the node agents ({\texttt{ComputeObs()}\xspace} and {\texttt{Schedule()}\xspace}), and {\texttt{PreemptionManager}\xspace} to support checkpointing through asynchronous preemption protocol. {\texttt{Scheduler}\xspace} provides an interface to load the checkpoint of the trained policy. It also supports a set of widely used scheduling policies, where users can configure one via script. {\texttt{NodeAgent}\xspace} is implemented on top of Slurm daemon ({\texttt{slurmd}\xspace}) with {\texttt{CheckpointManager}\xspace}, {\texttt{Profiler}\xspace} built on top of PyTorch profiler~\cite{PyTorchProf}, and {\texttt{ProcessManager}\xspace} for initializing a node process group used for synchronization in distributed DL training.

\begin{figure}[t]
    \centering
    \includegraphics[width=.99\columnwidth]{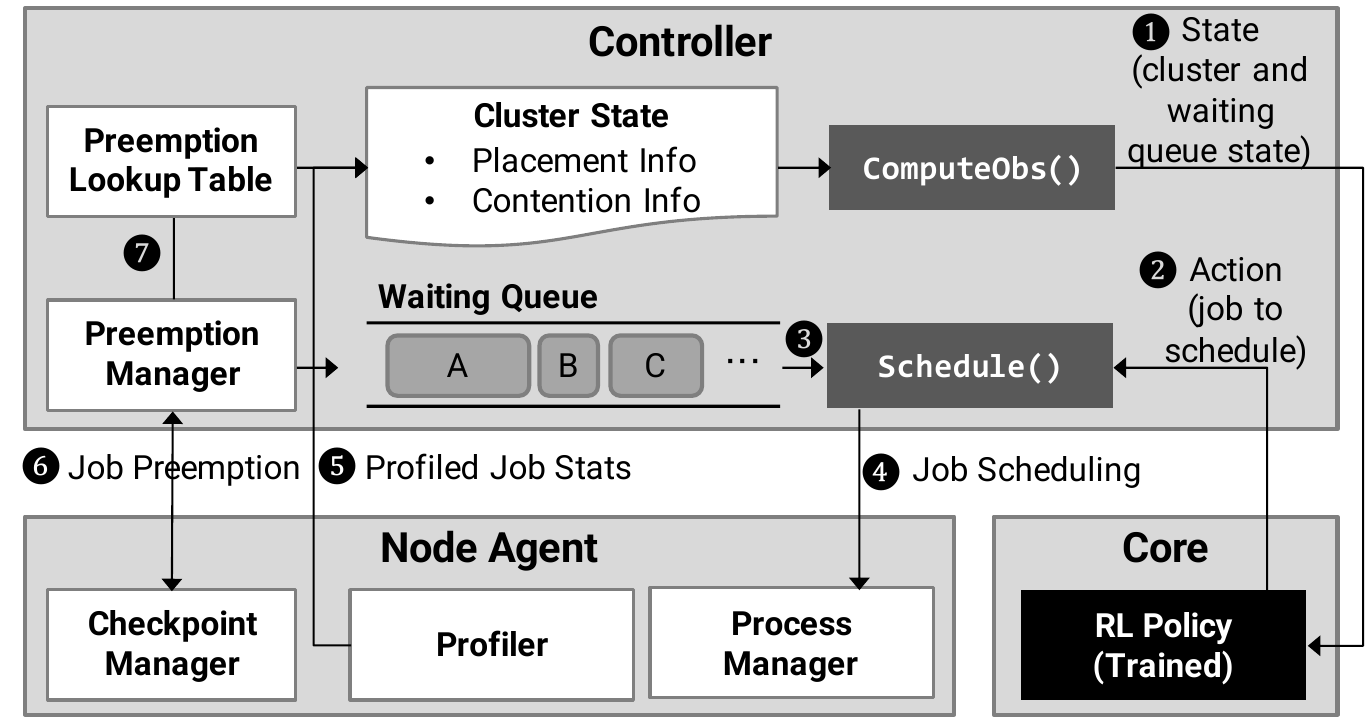}
    \caption{Detailed workflow of {\texttt{Scheduler}\xspace}.}
    \label{fig:scheduling}
    % \vspace{-1mm}
\end{figure}

\autoref{fig:scheduling} illustrates the in-depth workflow of {\texttt{Scheduler}\xspace}. The trained policy is plugged into {\texttt{Core}\xspace} and deployed on the GPU cluster. {\texttt{Core}\xspace} executes round-based scheduling with a predefined round duration knob. For every round, {\texttt{ComputeObs()}\xspace} encodes the cluster and the waiting queue state as a state and sends it to {\texttt{Core}\xspace}(\ballnumber{1}). Following the scheduling decision produced by the policy (\ballnumber{2}), {\texttt{Controller}\xspace} dequeues the target jobs to schedule from the waiting queue (\ballnumber{3}). Then {\texttt{Schedule()}\xspace} spawns {\texttt{NodeAgent}\xspace} for the job (one per each node allocated to the job), and submits the job training script (\ballnumber{4}). {\texttt{NodeAgent}\xspace} builds the process group for the job and delivers the profiled job stats to {\texttt{Controller}\xspace} after executing an online profiling for predefined iterations (\ballnumber{5}). Monitoring the contention sensitivity of jobs from the profiles, {\texttt{Controller}\xspace} sends a preemption signal to node agents with jobs with higher contention sensitivity than a predefined knob. {\texttt{NodeAgent}\xspace} waits for {\texttt{CheckpointManager}\xspace} to save the checkpoint of the job to distributed file system (DFS)-based job checkpoint store and returns its status to {\texttt{PreemptionManager}\xspace} asynchronously (\ballnumber{6}). To support re-scheduling, {\texttt{PreemptionManager}\xspace} maintains a preemption lookup table. Thus, whether a job has been preempted can be identified by querying the job identifier on the lookup table, and the scheduler loads from the checkpoint store to resume from the latest checkpoint (\ballnumber{7}).

\section{Evaluation}
\label{sec:eval}

We evaluate the performance of our approach using the two benchmark policies, \rlpolicy and \rlpolicyhybrid (Section \ref{sec:rl}).

\noindent
\textbf{Environment.} Experiment conducts on a homogeneous cluster of four nodes, each equipped with eight NVIDIA TITAN XP GPUs connected over 16GBps PCIe Gen3. Each node has one Mellanox MT27700 family ConnectX-4 NIC and is interconnected with 10Gbps Ethernet and 40Gbps InfiniBand. 

We believe our evaluation environment of 32 GPUs covers a large section of important use cases: Microsoft reports that 93.7\% of all jobs submitted to their internal DGX-2 clusters required at most 32 GPUs in the second half of 2021~\cite{shah2023taccl}. Also, despite the moderate scale of the evaluation environment, we note that our design can be applied to large-scale clusters without any limitation.

\noindent
\textbf{Traces.} 
To evaluate different scenarios with varying degrees of communication intensities, we generate four traces with different communication intensities according to the categorization in Section \ref{sec:motiv}. 
\footnote{Normal, heavy, medium, and low communication intensive traces follow the model distribution of GNN:IMG:DLRM:LM:FSDP:MoE=1:1:1:1:1:1, 1:1:1:1:4:4, 1:1:4:4:1:1, and 4:4:1:1:1:1, respectively.}
Each trace contains ten job sets each with 256 randomly shuffled jobs.
Each job is randomly assigned a GPU demand of up to 32, and given a total number of samples to train, which is configured to set its expected training time in an isolated cluster as one hour. The jobs' configurations (e.g., batch size, parameter size, number of layers, and gradient accumulation steps) are initialized randomly to evaluate extensibility to unseen cases.  

\begin{figure}[t!]
    \centering
    \includegraphics[width=.95\columnwidth]{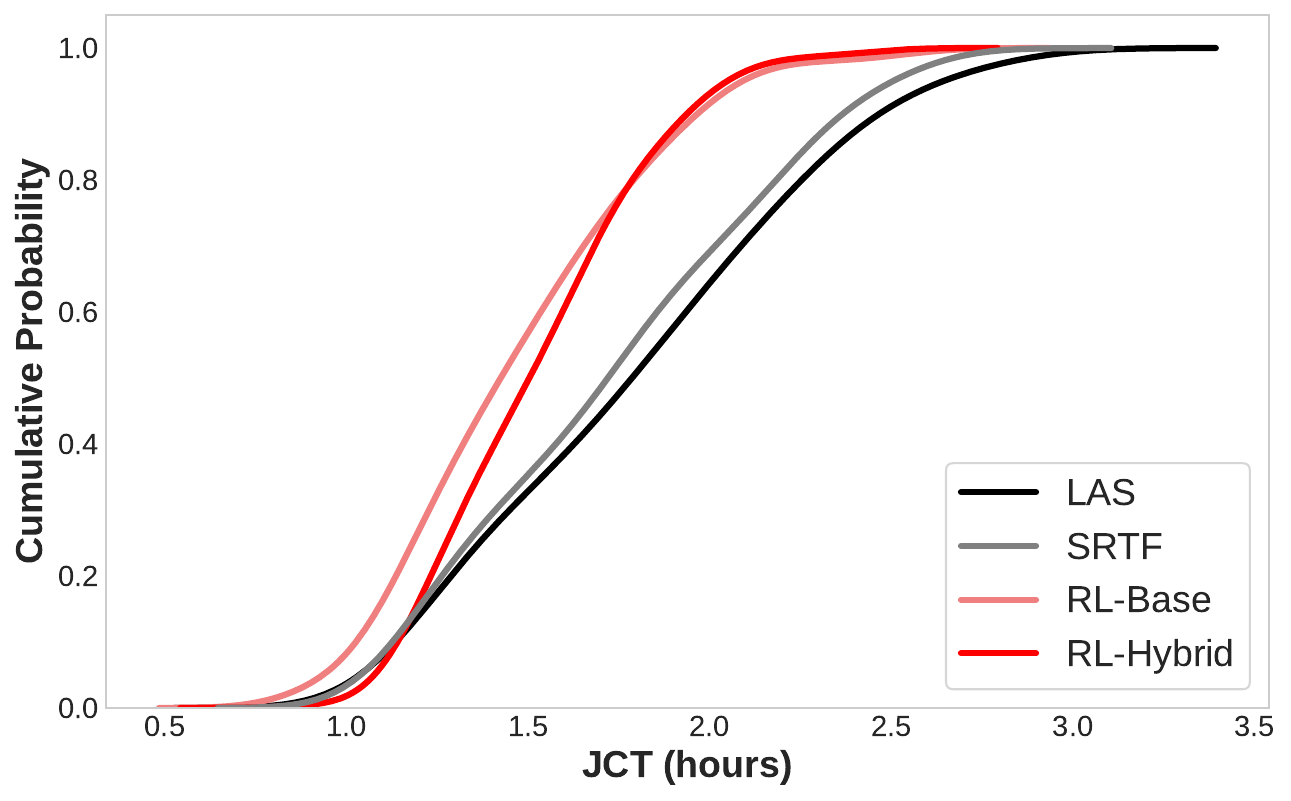}
    \captionsetup{justification=centering}
    \caption{CDF of JCT. (Trace: Normal, Branch: B)}
    \label{fig:jct-cdf}
    % \vspace{-5mm}
\end{figure}

\begin{figure}[t!]
    \centering
    \includegraphics[width=.99\columnwidth]{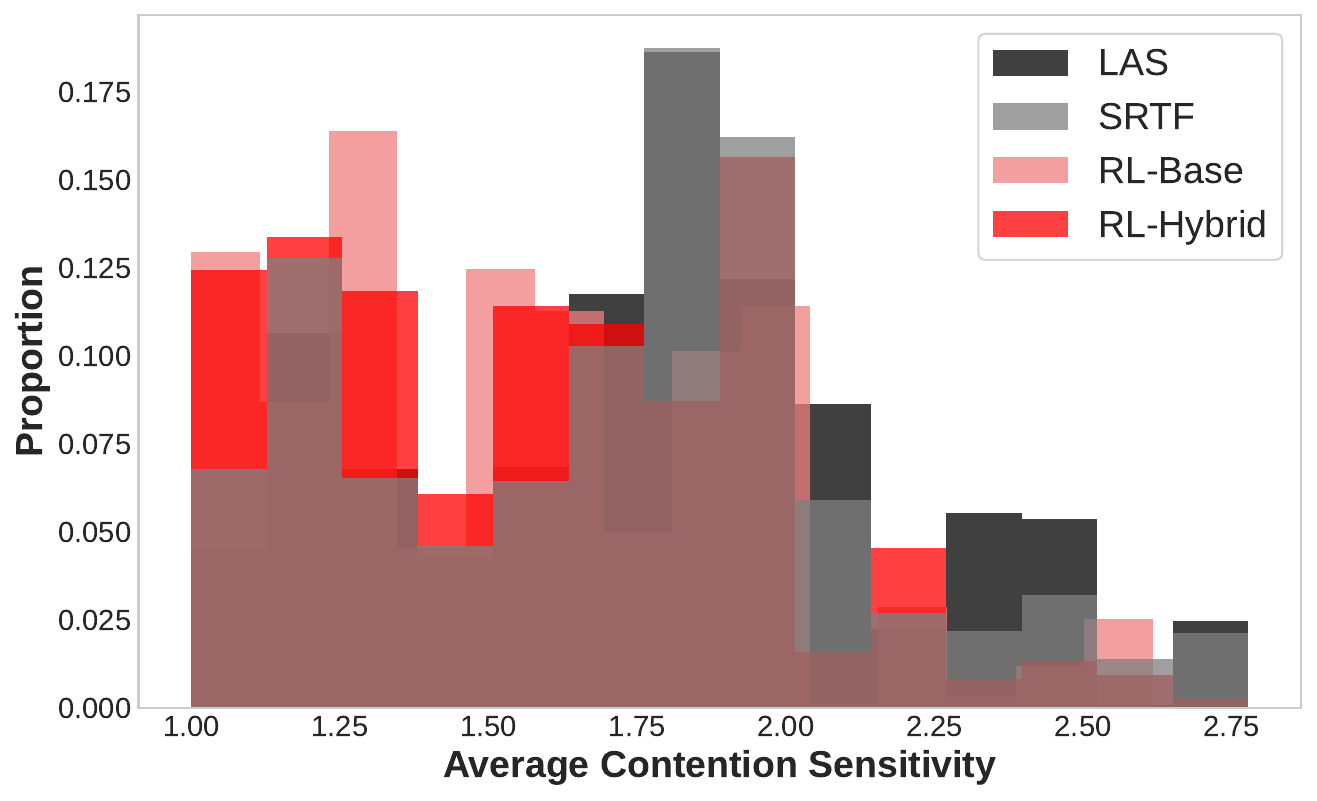}
    \captionsetup{justification=centering}
    \caption{Proportion of average contention sensitivity. \\(Trace: Normal, Branch: B)}
    \label{fig:avg-cs}
    % \vspace{-5mm}
\end{figure}

\begin{figure}[t!]
    \centering
    \includegraphics[width=.99\columnwidth]{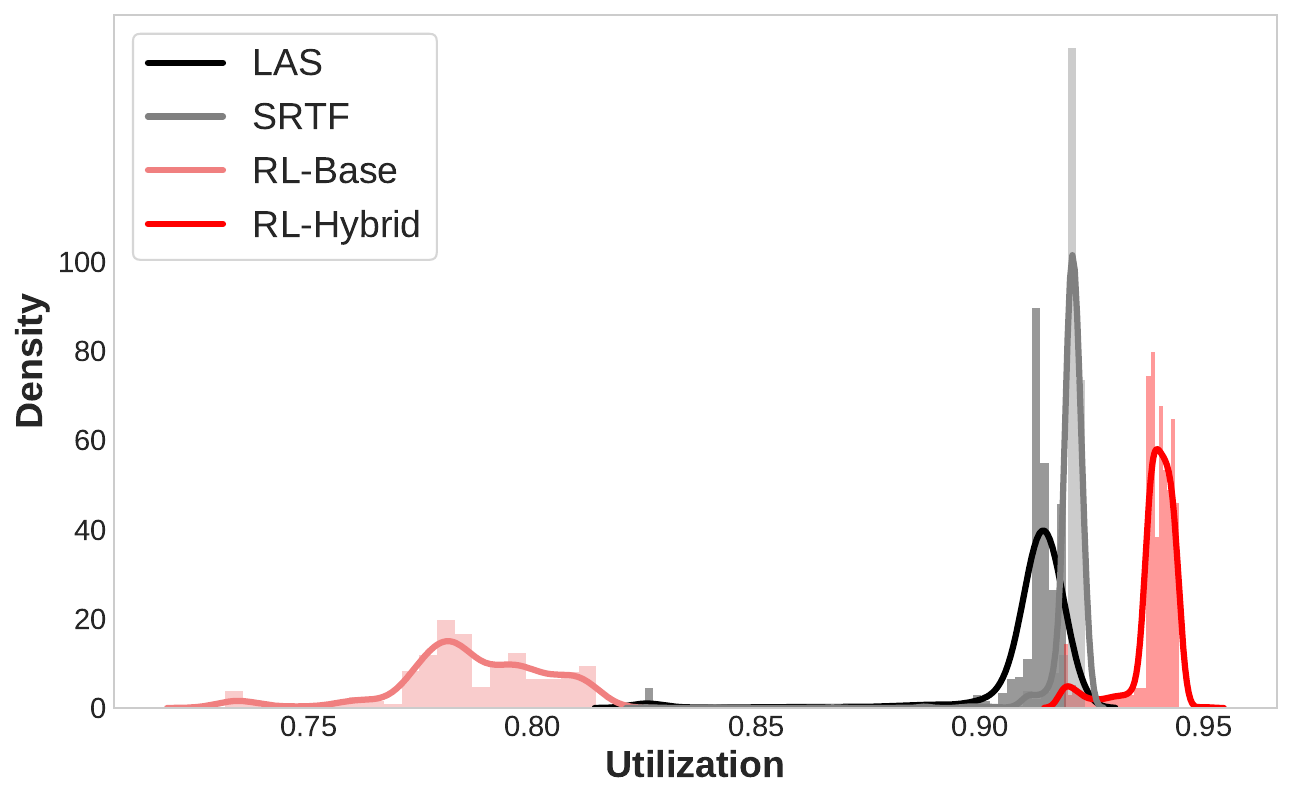}
    \captionsetup{justification=centering}
    \caption{KDE of utilization. (Trace: Normal, Branch: B)}
    \label{fig:util-kde}
    \vspace{-1mm}
\end{figure}

\begin{figure}[t!]
    \subfloat[Normal communication traces]
    {\includegraphics[width=.49\columnwidth]{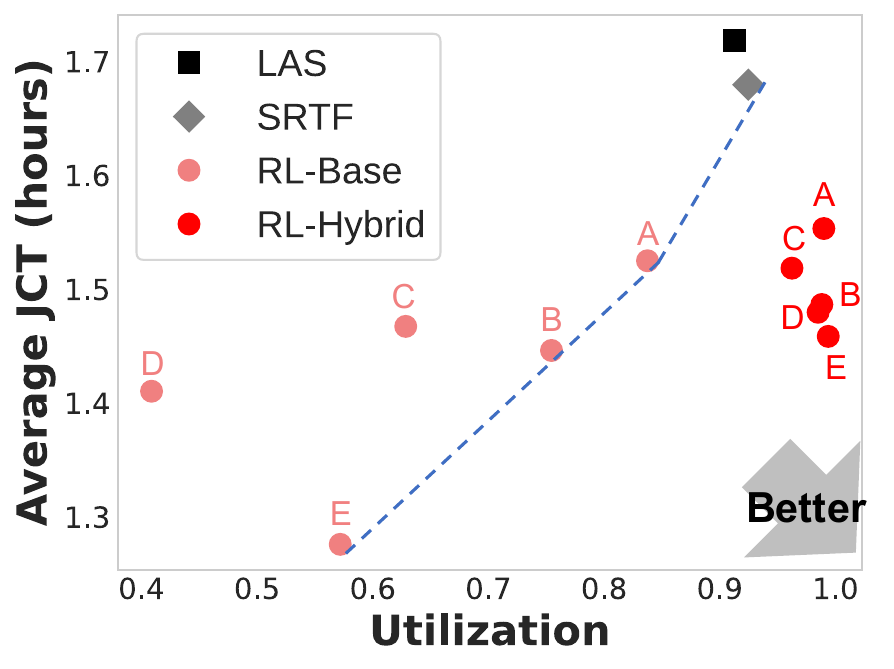}\label{fig:simul-eval-1}}
    \subfloat[Heavy communication traces]
    {\includegraphics[width=.49\columnwidth]{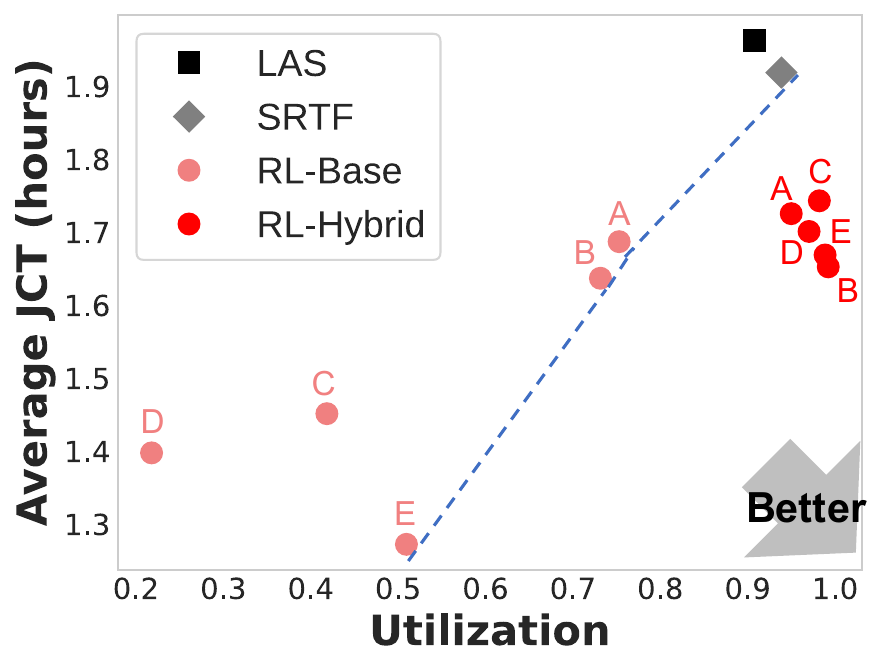}\label{fig:simul-eval-2}}
    \hfill
    \subfloat[Medium communication traces]
    {\includegraphics[width=.49\columnwidth]{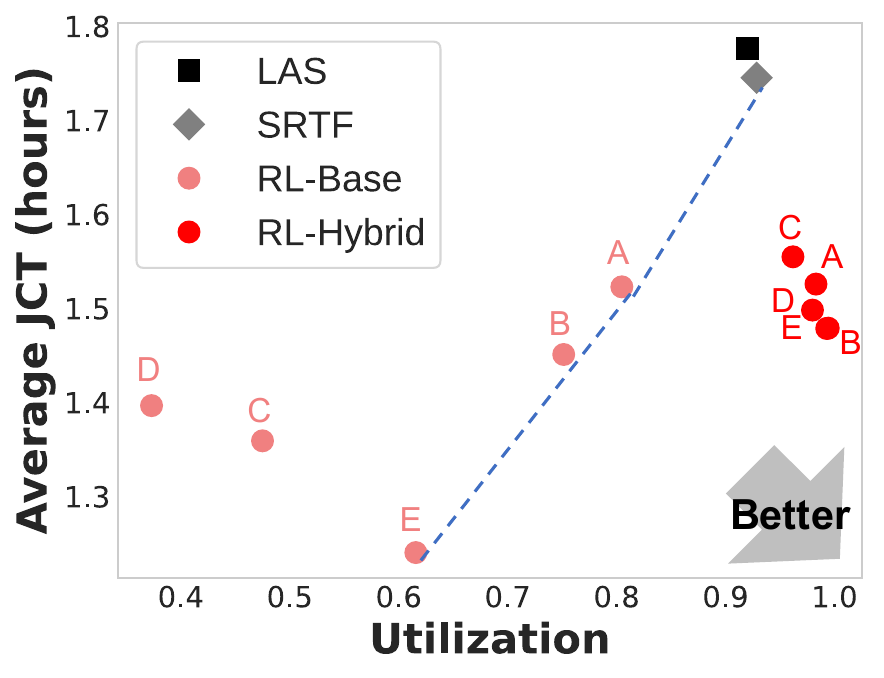}\label{fig:simul-eval-3}}
    \subfloat[Low communication traces]
    {\includegraphics[width=.49\columnwidth]{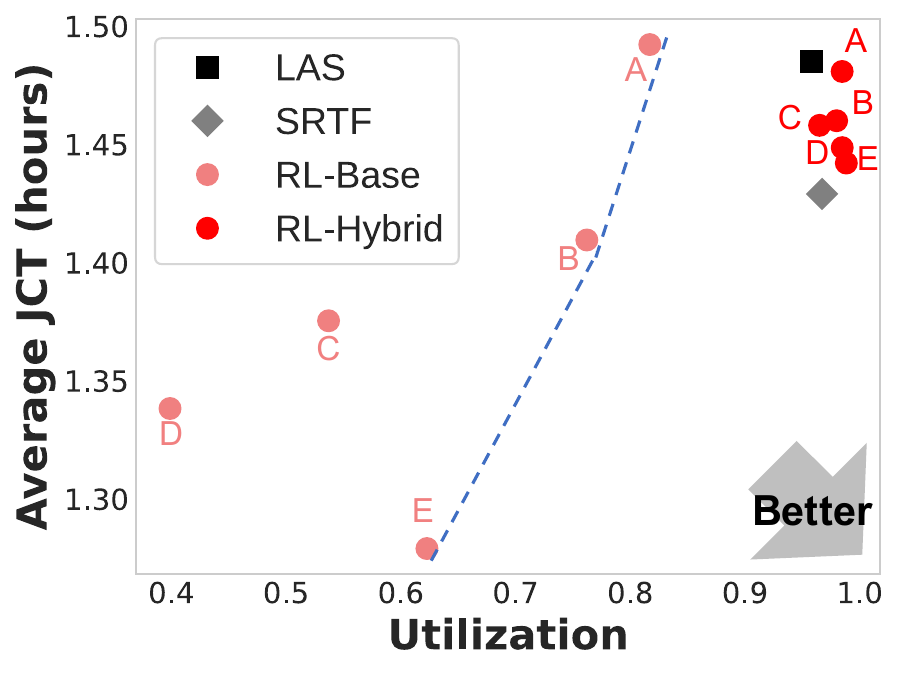}\label{fig:simul-eval-4}}
    \caption{Performance comparison for varying traces. The light red and red colored dots are different branches of \rlpolicy and \rlpolicyhybrid, respectively. Utilization denotes the ratio of used GPUs to total GPUs in the cluster.}
    \label{fig:simul-eval}
    % \vspace{-1mm}
\end{figure}

\noindent
\textbf{Branches of \rlpolicy and \rlpolicyhybrid.} We use five branches of \rlpolicy and \rlpolicyhybrid each by sweeping the range of the weights in the reward ($\wone$, $\wtwo$) (Section \ref{sec:rl}) to demonstrate trade-off relationship between average contention sensitivity and utilization. \footnote{We train each branch for 10 sets of jobs, each with 20 episodes that each contains 256 jobs.}  
In \autoref{fig:jct-cdf} to \autoref{fig:simul-eval},
A, B, C, D, and E correspond the branch with (0.3, 0.7), (0.4, 0.6), (0.5, 0.5), (0.6, 0.4), and (0.7, 0.3) of the weights, respectively. 

\noindent
\textbf{Results.} 
\autoref{fig:jct-cdf} shows the cumulative distribution function (CDF) of JCT. 
% avg
Comparing average JCT to SRTF and LAS, \rlpolicy (\rlpolicyhybrid) achieves 15.4\% (12.1\%) and 18.2\% (15.1\%) reduction, respectively.
% avg interpret
The large improvement in average JCT is the result of reduced cluster-wide contention sensitivity. 
\autoref{fig:avg-cs} shows the proportion of average contention sensitivity (i.e., average of contention sensitivities that jobs experience during training). \rlpolicy and \rlpolicyhybrid show a concentration of average contention sensitivity at a lower degree than SRTF and LAS. 
Clearly, this demonstrates that scheduling decisions of \rlpolicy and \rlpolicyhybrid successfully reduce cluster-wide average contention sensitivity which leads to an eventual large reduction in cluster-wide average JCT.
% p90
For p90 JCT, \rlpolicy and \rlpolicyhybrid equally achieve 16.4\% and 20.7\% reduction compared to SRTF and LAS, respectively.
Even though \rlpolicyhybrid shows the smaller degree of average JCT reduction ($\approx3\%$) than \rlpolicy, \rlpolicyhybrid trades such cost to a large improvement in cluster-wide utilization. \autoref{fig:util-kde} shows the kernel density estimation (KDE) of utilization. \rlpolicyhybrid displays the highest utilization, emphasizing the preferable trade-off between average JCT and utilization.
The difference mainly stems from their policy discrepancies; when the trained policy decides $\emptyset$, i.e., not to schedule, \rlpolicyhybrid tries to improve utilization by multiplexing to greedy scheduling.

% interpret tail JCT
The reduced tail JCT demonstrates that scheduling with RL effectively curbs scheduling decisions that trigger the worst contention situations.  

\autoref{fig:simul-eval} shows the results according to traces with varying degrees of communication intensities. In normal, high, and medium communication traces, we observe a trade-off relationship between average JCT and utilization among \rlpolicy, SRTF, and LAS (blue dashed lines). \rlpolicy's A and B trained with average contention sensitivity term weight ($w_{1}$) 0.3 and 0.4 achieve relatively balanced average JCT and utilization. On the other hand, \rlpolicy's C and D fail to reach the preferable trade-off. \rlpolicy's E trained with the highest penalization of the average contention sensitivity achieves the best average JCT at the cost of much lower utilization compared with SRTF and LAS which opt for maximizing utilization while not taking contention into account. \rlpolicyhybrid moves the original dots of \rlpolicy to the higher right end of the graph. This means that when cluster job distribution follows normal, high, or medium communication intensity conditions, adhering to the action of the trained RL policy in common cases while conforming to the decisions produced by the rule-based safety condition to prevent low utilization in rare cases leads to a large improvement in utilization at the expense of a relatively small increase in average JCT. 
Conversely, in low communication traces, we observe that SRTF and LAS perform similarly to \rlpolicyhybrid because the relatively low level of contention sensitivities of jobs makes the impact of network contention negligible. 

% \begin{figure}[]
%     \centering
%     \includegraphics[width=0.99\columnwidth]{figures/interpret.pdf}
%     \caption{Example snapshot during scheduling with \texttt{RL-Base}.}
%     \label{fig:interpret}
% \end{figure}

% Scheduling decisions made by \texttt{RL-Base} and \texttt{RL-Hybrid} provides interpretability through cluster state snapshots after each scheduling round. \autoref{fig:interpret} is an example snapshot we take during scheduling with \texttt{RL-Base} on normal communication traces.           
\section{Related Work}
\label{sec:relwork}

\noindent
\textbf{GPU cluster schedulers for DL.} 
GPU clusters employ a scheduler for efficient allocation of cluster resources. The common goals of the schedulers include reducing average JCT~\cite{optimus,tiresias,gavel,afs,pollux,muri}, improving utilization~\cite{gandiva,tiresias,antman,synergy}, or providing fairness~\cite{themis,gandivafair}. 
Optimus~\cite{optimus} estimates training speed and assigns more resources to jobs with higher marginal gain. 
Gandiva~\cite{gandiva} employs job time-slicing to provide early feedback and executes a greedy approach for job migration and packing.
Tiresias~\cite{tiresias} introduces \textit{least attained service} metric to prevent starvation of late arriving jobs.
Themis~\cite{themis} introduces \textit{finish time fairness} and promotes fairness between jobs.
Gandiva$_{\mathrm fair}$ guarantees fairness between users through allocating proportional resource shares.
Antman~\cite{antman} segments jobs into prioritized and opportunistic jobs to apply different scheduling policies.
Gavel~\cite{gavel} proposes a policy-agnostic scheduling mechanism and schedules based on throughput metric that is comparable across heterogeneous accelerator types.
AFS~\cite{afs} opts for balance between the \textit{shortest job first} policy and resource efficiency when allocating idle resources to jobs. 
Pollux~\cite{pollux} co-optimizes system throughput and job-level performance metrics.
Synergy~\cite{synergy} allocates host resources (e.g., CPU and DRAM) disproportionately according to jobs' sensitivity to resources.
Muri~\cite{muri} enables multiple jobs to be packed on the same set of resources by interleaving different stages of different jobs.
However, our work presents the first approach that identifies the network contention sensitivity of jobs and optimizes the cluster objectives by efficiently mitigating network contention.

\noindent
\textbf{RL for systems.}
Using RL for cluster scheduling allows rapid adaptation by learning to schedule DL training jobs that experience different levels of network contention sensitivities, where the distribution of jobs with different model, GPU demand, and placement combinations changes constantly.
The potential of the RL-based approach in high adaptivity to constantly changing or even unseen conditions has been already shown in various domains. In TCP congestion control (CC), PCC-RL~\cite{pcc-rl} shows that RL-based TCP CC algorithm can successfully learn to distinguish different types of network losses that hand-optimized TCP CUBIC~\cite{tcp-cubic} cannot capture. In adaptive video streaming, Pensieve~\cite{pensieve} demonstrates that RL-based adaptive bitrate selection for video streaming can improve quality-of-experience (QoE) such as reduced frame stalling while improving bandwidth utilization. Merina~\cite{merina} shows that using meta-RL allows fast adaptation in unseen throughput dynamics, further improving QoE across a wide range of network throughput patterns.
Our work applies RL in resource management system, and present the first approach to propose network contention-aware GPU cluster sheduling with RL.         
\section{Conclusion}
\label{sec:conc}

We present a novel design that translates the network contention problem in GPU cluster scheduling into an RL problem. Our RL formulation trains the scheduling policies to efficiently tackle the challenge of continuously changing contention sensitivities of jobs in GPU clusters. We build an end-to-end system that can train scheduling policies with RL and deploy on GPU clusters. Our evaluation show that RL-based scheduling policies achieve reduced average and tail JCT by up to 18.2\% and 20.7\% compared to the widely used LAS and SRTF scheduling policies, and allows preferable trade-off of large improvement of utilization with small cost in average JCT. Our work is open-sourced at \url{https://github.com/gajagajago/deepshare} for future research in RL-based GPU cluster scheduling.           

\bibliographystyle{IEEEtran}
\bibliography{IEEEabrv,deepshare}

\begin{thebibliography}{10}
\providecommand{\url}[1]{#1}
\csname url@samestyle\endcsname
\providecommand{\newblock}{\relax}
\providecommand{\bibinfo}[2]{#2}
\providecommand{\BIBentrySTDinterwordspacing}{\spaceskip=0pt\relax}
\providecommand{\BIBentryALTinterwordstretchfactor}{4}
\providecommand{\BIBentryALTinterwordspacing}{\spaceskip=\fontdimen2\font plus
\BIBentryALTinterwordstretchfactor\fontdimen3\font minus
  \fontdimen4\font\relax}
\providecommand{\BIBforeignlanguage}[2]{{%
\expandafter\ifx\csname l@#1\endcsname\relax
\typeout{** WARNING: IEEEtran.bst: No hyphenation pattern has been}%
\typeout{** loaded for the language `#1'. Using the pattern for}%
\typeout{** the default language instead.}%
\else
\language=\csname l@#1\endcsname
\fi
#2}}
\providecommand{\BIBdecl}{\relax}
\BIBdecl

\bibitem{mlaas}
Q.~Weng, W.~Xiao, Y.~Yu, W.~Wang, C.~Wang, J.~He, Y.~Li, L.~Zhang, W.~Lin, and
  Y.~Ding, ``{MLaaS in the Wild: Workload Analysis and Scheduling in
  Large-Scale Heterogeneous GPU Clusters},'' in \emph{19th USENIX Symposium on
  Networked Systems Design and Implementation (NSDI 22)}, 2022, pp. 945--960.

\bibitem{philly}
M.~Jeon, S.~Venkataraman, A.~Phanishayee, J.~Qian, W.~Xiao, and F.~Yang,
  ``{Analysis of Large-Scale Multi-Tenant GPU Clusters for DNN Training
  Workloads},'' in \emph{2019 USENIX Annual Technical Conference (USENIX ATC
  19)}, 2019, pp. 947--960.

\bibitem{fairscale-fsdp}
\BIBentryALTinterwordspacing
M.~Ott, S.~Shleifer, M.~Xu, P.~Goyal, Q.~D. Duval, and V.~Caggiano. (2021)
  {Fully Sharded Data Parallel: faster AI training with fewer GPUs}. [Online].
  Available: \url{https://engineering.fb.com/2021/07/15/open-source/fsdp/}
\BIBentrySTDinterwordspacing

\bibitem{pytorch-fsdp}
\BIBentryALTinterwordspacing
Y.~Zhao, R.~Varma, C.-C. Huang, S.~Li, X.~Min, and A.~D. Desmaison. (2022)
  {Introducing PyTorch Fully Sharded Data Parallel (FSDP) API}. [Online].
  Available:
  \url{https://pytorch.org/blog/introducing-pytorch-fully-sharded-data-parallel-api/}
\BIBentrySTDinterwordspacing

\bibitem{zero}
S.~Rajbhandari, J.~Rasley, O.~Ruwase, and Y.~He, ``Zero: Memory optimizations
  toward training trillion parameter models,'' in \emph{SC20: International
  Conference for High Performance Computing, Networking, Storage and
  Analysis}.\hskip 1em plus 0.5em minus 0.4em\relax IEEE, 2020, pp. 1--16.

\bibitem{moe}
N.~Shazeer, A.~Mirhoseini, K.~Maziarz, A.~Davis, Q.~Le, G.~Hinton, and J.~Dean,
  ``Outrageously large neural networks: The sparsely-gated mixture-of-experts
  layer,'' \emph{arXiv preprint arXiv:1701.06538}, 2017.

\bibitem{muri}
Y.~Zhao, Y.~Liu, Y.~Peng, Y.~Zhu, X.~Liu, and X.~Jin, ``Multi-resource
  interleaving for deep learning training,'' in \emph{Proceedings of the ACM
  SIGCOMM 2022 Conference}, 2022, pp. 428--440.

\bibitem{rajasekaran2022congestion}
S.~Rajasekaran, M.~Ghobadi, G.~Kumar, and A.~Akella, ``Congestion control in
  machine learning clusters,'' in \emph{Proceedings of the 21st ACM Workshop on
  Hot Topics in Networks}, 2022, pp. 235--242.

\bibitem{tiresias}
J.~Gu, M.~Chowdhury, K.~G. Shin, Y.~Zhu, M.~Jeon, J.~Qian, H.~Liu, and C.~Guo,
  ``{Tiresias: A GPU cluster manager for distributed deep learning},'' in
  \emph{16th USENIX Symposium on Networked Systems Design and Implementation
  (NSDI 19)}, 2019, pp. 485--500.

\bibitem{las}
M.~Nuyens and A.~Wierman, ``The foreground--background queue: a survey,''
  \emph{Performance evaluation}, vol.~65, no. 3-4, pp. 286--307, 2008.

\bibitem{srtf}
M.~E. Crovella, R.~Frangioso, and M.~Harchol-Balter, ``Connection scheduling in
  web servers,'' Citeseer, Tech. Rep., 1999.

\bibitem{shinde2018review}
P.~P. Shinde and S.~Shah, ``A review of machine learning and deep learning
  applications,'' in \emph{2018 Fourth international conference on computing
  communication control and automation (ICCUBEA)}.\hskip 1em plus 0.5em minus
  0.4em\relax IEEE, 2018, pp. 1--6.

\bibitem{synergy}
J.~Mohan, A.~Phanishayee, J.~Kulkarni, and V.~Chidambaram, ``{Looking beyond
  GPUs for DNN scheduling on Multi-Tenant clusters},'' in \emph{16th USENIX
  Symposium on Operating Systems Design and Implementation (OSDI 22)}, 2022,
  pp. 579--596.

\bibitem{optimus}
Y.~Peng, Y.~Bao, Y.~Chen, C.~Wu, and C.~Guo, ``Optimus: an efficient dynamic
  resource scheduler for deep learning clusters,'' in \emph{Proceedings of the
  Thirteenth EuroSys Conference}, 2018, pp. 1--14.

\bibitem{gavel}
D.~Narayanan, K.~Santhanam, F.~Kazhamiaka, A.~Phanishayee, and M.~Zaharia,
  ``{Heterogeneity-Aware cluster scheduling policies for deep learning
  workloads},'' in \emph{14th USENIX Symposium on Operating Systems Design and
  Implementation (OSDI 20)}, 2020, pp. 481--498.

\bibitem{afs}
C.~Hwang, T.~Kim, S.~Kim, J.~Shin, and K.~Park, ``Elastic resource sharing for
  distributed deep learning,'' in \emph{18th USENIX Symposium on Networked
  Systems Design and Implementation (NSDI 21)}, 2021, pp. 721--739.

\bibitem{pollux}
A.~Qiao, S.~K. Choe, S.~J. Subramanya, W.~Neiswanger, Q.~Ho, H.~Zhang, G.~R.
  Ganger, and E.~P. Xing, ``Pollux: Co-adaptive cluster scheduling for
  goodput-optimized deep learning,'' in \emph{15th USENIX Symposium on
  Operating Systems Design and Implementation (OSDI 21)}, 2021.

\bibitem{gandiva}
W.~Xiao, R.~Bhardwaj, R.~Ramjee, M.~Sivathanu, N.~Kwatra, Z.~Han, P.~Patel,
  X.~Peng, H.~Zhao, Q.~Zhang \emph{et~al.}, ``Gandiva: Introspective cluster
  scheduling for deep learning,'' in \emph{13th USENIX Symposium on Operating
  Systems Design and Implementation (OSDI 18)}, 2018, pp. 595--610.

\bibitem{antman}
W.~Xiao, S.~Ren, Y.~Li, Y.~Zhang, P.~Hou, Z.~Li, Y.~Feng, W.~Lin, and Y.~Jia,
  ``{AntMan: Dynamic Scaling on GPU Clusters for Deep Learning},'' in
  \emph{14th USENIX Symposium on Operating Systems Design and Implementation
  (OSDI 20)}, 2020, pp. 533--548.

\bibitem{sutton2018reinforcement}
R.~S. Sutton and A.~G. Barto, \emph{Reinforcement learning: An
  introduction}.\hskip 1em plus 0.5em minus 0.4em\relax MIT press, 2018.

\bibitem{deeprm}
H.~Mao, M.~Alizadeh, I.~Menache, and S.~Kandula, ``Resource management with
  deep reinforcement learning,'' in \emph{Proceedings of the 15th ACM workshop
  on hot topics in networks}, 2016, pp. 50--56.

\bibitem{mnih2016asynchronous}
V.~Mnih, A.~P. Badia, M.~Mirza, A.~Graves, T.~Lillicrap, T.~Harley, D.~Silver,
  and K.~Kavukcuoglu, ``Asynchronous methods for deep reinforcement learning,''
  in \emph{International conference on machine learning}.\hskip 1em plus 0.5em
  minus 0.4em\relax PMLR, 2016, pp. 1928--1937.

\bibitem{mnih2013playing}
V.~Mnih, K.~Kavukcuoglu, D.~Silver, A.~Graves, I.~Antonoglou, D.~Wierstra, and
  M.~Riedmiller, ``Playing atari with deep reinforcement learning,''
  \emph{arXiv preprint arXiv:1312.5602}, 2013.

\bibitem{mnih2015human}
V.~Mnih, K.~Kavukcuoglu, D.~Silver, A.~A. Rusu, J.~Veness, M.~G. Bellemare,
  A.~Graves, M.~Riedmiller, A.~K. Fidjeland, G.~Ostrovski \emph{et~al.},
  ``Human-level control through deep reinforcement learning,'' \emph{nature},
  vol. 518, no. 7540, pp. 529--533, 2015.

\bibitem{schulman2015trust}
J.~Schulman, S.~Levine, P.~Abbeel, M.~Jordan, and P.~Moritz, ``Trust region
  policy optimization,'' in \emph{International conference on machine
  learning}.\hskip 1em plus 0.5em minus 0.4em\relax PMLR, 2015, pp. 1889--1897.

\bibitem{silver2016mastering}
D.~Silver, A.~Huang, C.~J. Maddison, A.~Guez, L.~Sifre, G.~Van Den~Driessche,
  J.~Schrittwieser, I.~Antonoglou, V.~Panneershelvam, M.~Lanctot \emph{et~al.},
  ``{Mastering the game of Go with deep neural networks and tree search},''
  \emph{nature}, vol. 529, no. 7587, pp. 484--489, 2016.

\bibitem{pcc-rl}
N.~Jay, N.~Rotman, B.~Godfrey, M.~Schapira, and A.~Tamar, ``A deep
  reinforcement learning perspective on internet congestion control,'' in
  \emph{International Conference on Machine Learning}.\hskip 1em plus 0.5em
  minus 0.4em\relax PMLR, 2019, pp. 3050--3059.

\bibitem{pensieve}
H.~Mao, R.~Netravali, and M.~Alizadeh, ``Neural adaptive video streaming with
  pensieve,'' in \emph{Proceedings of the conference of the ACM special
  interest group on data communication}, 2017, pp. 197--210.

\bibitem{comyco}
T.~Huang, C.~Zhou, X.~Yao, R.-X. Zhang, C.~Wu, B.~Yu, and L.~Sun,
  ``Quality-aware neural adaptive video streaming with lifelong imitation
  learning,'' \emph{IEEE Journal on Selected Areas in Communications}, vol.~38,
  no.~10, pp. 2324--2342, 2020.

\bibitem{merina}
N.~Kan, Y.~Jiang, C.~Li, W.~Dai, J.~Zou, and H.~Xiong, ``{Improving
  Generalization for Neural Adaptive Video Streaming via Meta Reinforcement
  Learning},'' in \emph{Proceedings of the 30th ACM International Conference on
  Multimedia}, 2022, pp. 3006--3016.

\bibitem{r3net}
J.~Fang, M.~Ellis, B.~Li, S.~Liu, Y.~Hosseinkashi, M.~Revow, A.~Sadovnikov,
  Z.~Liu, P.~Cheng, S.~Ashok \emph{et~al.}, ``Reinforcement learning for
  bandwidth estimation and congestion control in real-time communications,''
  \emph{arXiv preprint arXiv:1912.02222}, 2019.

\bibitem{concerto}
A.~Zhou, H.~Zhang, G.~Su, L.~Wu, R.~Ma, Z.~Meng, X.~Zhang, X.~Xie, H.~Ma, and
  X.~Chen, ``Learning to coordinate video codec with transport protocol for
  mobile video telephony,'' in \emph{The 25th Annual International Conference
  on Mobile Computing and Networking}, 2019, pp. 1--16.

\bibitem{onrl}
H.~Zhang, A.~Zhou, J.~Lu, R.~Ma, Y.~Hu, C.~Li, X.~Zhang, H.~Ma, and X.~Chen,
  ``{OnRL: improving mobile video telephony via online reinforcement
  learning},'' in \emph{Proceedings of the 26th Annual International Conference
  on Mobile Computing and Networking}, 2020.

\bibitem{loki}
H.~Zhang, A.~Zhou, Y.~Hu, C.~Li, G.~Wang, X.~Zhang, H.~Ma, L.~Wu, A.~Chen, and
  C.~Wu, ``Loki: improving long tail performance of learning-based real-time
  video adaptation by fusing rule-based models,'' in \emph{Proceedings of the
  26th Annual International Conference on Mobile Computing and Networking},
  2021.

\bibitem{hrcc}
B.~Wang, Y.~Zhang, S.~Qian, Z.~Pan, and Y.~Xie, ``A hybrid receiver-side
  congestion control scheme for web real-time communication,'' in
  \emph{Proceedings of the 12th ACM Multimedia Systems Conference}, 2021, pp.
  332--338.

\bibitem{sibyl}
G.~Singh, R.~Nadig, J.~Park, R.~Bera, N.~Hajinazar, D.~Novo, J.~G{\'o}mez-Luna,
  S.~Stuijk, H.~Corporaal, and O.~Mutlu, ``Sibyl: Adaptive and extensible data
  placement in hybrid storage systems using online reinforcement learning,'' in
  \emph{Proceedings of the 49th Annual International Symposium on Computer
  Architecture}, 2022, pp. 320--336.

\bibitem{deepplace-apsys}
S.~Mitra, S.~S. Mondal, N.~Sheoran, N.~Dhake, R.~Nehra, and R.~Simha,
  ``Deepplace: Learning to place applications in multi-tenant clusters,'' in
  \emph{Proceedings of the 10th ACM SIGOPS Asia-Pacific Workshop on Systems},
  2019, pp. 61--68.

\bibitem{deepplace-aaai}
S.~S. Mondal, N.~Sheoran, and S.~Mitra, ``Scheduling of time-varying workloads
  using reinforcement learning,'' in \emph{Proceedings of the AAAI Conference
  on Artificial Intelligence}, vol.~35, no.~10, 2021, pp. 9000--9008.

\bibitem{graphsage}
W.~Hamilton, Z.~Ying, and J.~Leskovec, ``Inductive representation learning on
  large graphs,'' \emph{Advances in neural information processing systems},
  vol.~30, 2017.

\bibitem{mobilenet}
A.~Howard, M.~Sandler, G.~Chu, L.-C. Chen, B.~Chen, M.~Tan, W.~Wang, Y.~Zhu,
  R.~Pang, V.~Vasudevan \emph{et~al.}, ``Searching for mobilenetv3,'' in
  \emph{Proceedings of the IEEE/CVF international conference on computer
  vision}, 2019, pp. 1314--1324.

\bibitem{dlrm}
M.~Naumov, D.~Mudigere, H.-J.~M. Shi, J.~Huang, N.~Sundaraman, J.~Park,
  X.~Wang, U.~Gupta, C.-J. Wu, A.~G. Azzolini \emph{et~al.}, ``Deep learning
  recommendation model for personalization and recommendation systems,''
  \emph{arXiv preprint arXiv:1906.00091}, 2019.

\bibitem{transformerxl}
Z.~Dai, Z.~Yang, Y.~Yang, J.~Carbonell, Q.~V. Le, and R.~Salakhutdinov,
  ``Transformer-xl: Attentive language models beyond a fixed-length context,''
  \emph{arXiv preprint arXiv:1901.02860}, 2019.

\bibitem{radford2019language}
A.~Radford, J.~Wu, R.~Child, D.~Luan, D.~Amodei, I.~Sutskever \emph{et~al.},
  ``Language models are unsupervised multitask learners,'' \emph{OpenAI blog},
  vol.~1, no.~8, p.~9, 2019.

\bibitem{yangdlrmkdd}
J.~A. Yang, J.~Park, S.~Sridharan, and P.~T.~P. Tang, ``Training deep learning
  recommendation model with quantized collective communications,'' in
  \emph{Conference on Knowledge Discovery and Data Mining (KDD)}, 2020.

\bibitem{vaswani2017attention}
A.~Vaswani, N.~Shazeer, N.~Parmar, J.~Uszkoreit, L.~Jones, A.~N. Gomez,
  {\L}.~Kaiser, and I.~Polosukhin, ``Attention is all you need,''
  \emph{Advances in neural information processing systems}, vol.~30, 2017.

\bibitem{orca}
S.~Abbasloo, C.-Y. Yen, and H.~J. Chao, ``Classic meets modern: A pragmatic
  learning-based congestion control for the internet,'' in \emph{Proceedings of
  the Annual conference of the ACM Special Interest Group on Data Communication
  on the applications, technologies, architectures, and protocols for computer
  communication}, 2020, pp. 632--647.

\bibitem{brockman2016openai}
G.~Brockman, V.~Cheung, L.~Pettersson, J.~Schneider, J.~Schulman, J.~Tang, and
  W.~Zaremba, ``Openai gym,'' \emph{arXiv preprint arXiv:1606.01540}, 2016.

\bibitem{stable-baselines3}
A.~Raffin, A.~Hill, A.~Gleave, A.~Kanervisto, M.~Ernestus, and N.~Dormann,
  ``Stable-baselines3: Reliable reinforcement learning implementations,''
  \emph{The Journal of Machine Learning Research}, vol.~22, no.~1, pp.
  12\,348--12\,355, 2021.

\bibitem{yoo2003slurm}
A.~B. Yoo, M.~A. Jette, and M.~Grondona, ``Slurm: Simple linux utility for
  resource management,'' in \emph{Job Scheduling Strategies for Parallel
  Processing: 9th International Workshop, JSSPP 2003, Seattle, WA, USA, June
  24, 2003. Revised Paper 9}.\hskip 1em plus 0.5em minus 0.4em\relax Springer,
  2003, pp. 44--60.

\bibitem{shvachko2010hadoop}
K.~Shvachko, H.~Kuang, S.~Radia, and R.~Chansler, ``The hadoop distributed file
  system,'' in \emph{2010 IEEE 26th symposium on mass storage systems and
  technologies (MSST)}.\hskip 1em plus 0.5em minus 0.4em\relax Ieee, 2010, pp.
  1--10.

\bibitem{PyTorchProf}
``{PyTorch Profiler},'' 2023,
  \url{https://pytorch.org/tutorials/recipes/recipes/profiler_recipe.html}.

\bibitem{shah2023taccl}
A.~Shah, V.~Chidambaram, M.~Cowan, S.~Maleki, M.~Musuvathi, T.~Mytkowicz,
  J.~Nelson, O.~Saarikivi, and R.~Singh, ``{TACCL: Guiding Collective Algorithm
  Synthesis using Communication Sketches},'' in \emph{20th USENIX Symposium on
  Networked Systems Design and Implementation (NSDI 23)}, 2023, pp. 593--612.

\bibitem{themis}
K.~Mahajan, A.~Balasubramanian, A.~Singhvi, S.~Venkataraman, A.~Akella,
  A.~Phanishayee, and S.~Chawla, ``{Themis: Fair and efficient GPU cluster
  scheduling},'' in \emph{17th USENIX Symposium on Networked Systems Design and
  Implementation (NSDI 20)}, 2020, pp. 289--304.

\bibitem{gandivafair}
S.~Chaudhary, R.~Ramjee, M.~Sivathanu, N.~Kwatra, and S.~Viswanatha,
  ``{Balancing efficiency and fairness in heterogeneous GPU clusters for deep
  learning},'' in \emph{Proceedings of the Fifteenth European Conference on
  Computer Systems}, 2020, pp. 1--16.

\bibitem{tcp-cubic}
S.~Ha, I.~Rhee, and L.~Xu, ``{CUBIC: a new TCP-friendly high-speed TCP
  variant},'' \emph{ACM SIGOPS operating systems review}, vol.~42, no.~5, pp.
  64--74, 2008.

\end{thebibliography}

\end{document}